\documentclass[runningheads]{llncs}
\usepackage{multirow}
\usepackage{bm}
\usepackage{booktabs}
\usepackage{graphicx}
\usepackage{comment}
\usepackage{amsmath,amssymb} % define this before the line numbering.
\usepackage{color}

\usepackage[width=122mm,left=12mm,paperwidth=146mm,height=193mm,top=12mm,paperheight=217mm]{geometry}

\begin{document}
% \renewcommand\thelinenumber{\color[rgb]{0.2,0.5,0.8}\normalfont\sffamily\scriptsize\arabic{linenumber}\color[rgb]{0,0,0}}
% \renewcommand\makeLineNumber {\hss\thelinenumber\ \hspace{6mm} \rlap{\hskip\textwidth\ \hspace{6.5mm}\thelinenumber}}
% \linenumbers
\pagestyle{plain}
\mainmatter
\def\ECCVSubNumber{4261}  % Insert your submission number here
\title{Dual-attention Guided Dropblock Module for
Weakly Supervised Object Localization} % Replace with your title

% INITIAL SUBMISSION
\begin{comment}
\titlerunning{ECCV-20 submission ID \ECCVSubNumber}
\authorrunning{ECCV-20 submission ID \ECCVSubNumber}
%\author{Junhui Yin, Siqing Zhang, Dongliang Chang, Zhanyu Ma$^{*}$, Jun Guo}
\author{Anonymous ECCV submission}
\institute{Paper ID \ECCVSubNumber}
\end{comment}
%******************

% CAMERA READY SUBMISSION
%\begin{comment}
%\titlerunning{Abbreviated paper title}
% If the paper title is too long for the running head, you can set
% an abbreviated paper title here
%
\author{Junhui Yin, Siqing Zhang, Dongliang Chang, Zhanyu Ma, Jun Guo}
\institute{
Beijing University of Posts and Telecommunications}
% First names are abbreviated in the running head.
% If there are more than two authors, 'et al.' is used.

%\end{comment}
%******************
\maketitle

\begin{abstract}
%Weakly Supervised Object Localization (WSOL) techniques
%remain a challenge when capturing the object location only using image-level labels, without the guidance of location %annotations.
Attention mechanisms is frequently used to learn the discriminative features for better feature representations. In this paper, we extend the attention mechanism to the task of weakly supervised object localization (WSOL) and propose the dual-attention guided dropblock module (DGDM), which aims at learning the informative and complementary visual patterns for WSOL. This module contains two key components, the channel attention guided dropout (CAGD) and the spatial attention guided dropblock (SAGD). To model channel interdependencies,
the CAGD ranks the channel attentions and treats the top-$k$ attentions with the largest magnitudes as the important ones.
It also keeps some low-valued elements to increase their value if they become important
during training.
The SAGD can efficiently remove the most discriminative information by erasing the contiguous regions of feature maps rather than individual pixels. This guides the model to capture the less discriminative parts for classification. Furthermore, it can also distinguish the foreground objects from the background regions to alleviate the attention misdirection. Experimental results demonstrate that the proposed method achieves new state-of-the-art localization performance.
\end{abstract}

\section{Introduction}
% no \IEEEPARstart
Weakly supervised object localization (WSOL) requires less detailed annotations to
identify the object location in a given image \cite{zhou2016learning} compared to the fully-supervised learning. WSOL is a challenging task since neural networks have access to only image-level labels (``cat" or ``no cat") that confirms the existence of the target
object, but not the guidance of the expensive bounding box annotations in an image.

To address the WSOL problem with convolutional neural networks (CNNs), people resort to a general method, \emph{e.g.}, generating Class Activation Mapping (CAM) \cite{zhou2016learning} for
performing the object localization.
Unfortunately, the CAM solely discovers a small part of target objects instead of the entire object, which leads to localization accuracy degradation \cite{choe2019attention}.

Different from CAM, which relies only on the most discriminative information, existing approaches have explored adversarial erasing \cite{zhang2018adversarial}, Hide-and-Seek (HaS) \cite{singh2017hide},  Attention-based Dropout Layer (ADL) \cite{choe2019attention}.
Specifically, the Adversarial Complementary Learning (ACoL) approach \cite{zhang2018adversarial} can efficiently locate different object regions and learn new and complementary parts belonging to the same
objects by two adversary classifiers in a weakly supervised manner.
HaS \cite{singh2017hide} hides the patches randomly, which encourages the network to seek the multiple relevant parts. ADL \cite{choe2019attention} hides the most
discriminative part from the model for pursuing the full object extents and then highlights the informative
parts to improve the recognition power of CNNs model.
In fact, similar to the pixel-based dropout, these techniques are not really the region-based dropout that can efficiently remove the information. This is because the drop mask of ADL is generated
by thresholding the pixel values on the feature map.
However, the neighbouring pixels are spatially correlated on the feature map. These adjacent
pixels share much of the same information. The pixel-based dropout discards the information on the convolutional feature map. However, the information are still passed on from the adjacent pixels that
are left active.

\begin{figure}[t]
\centering
\includegraphics[height=2.5cm]{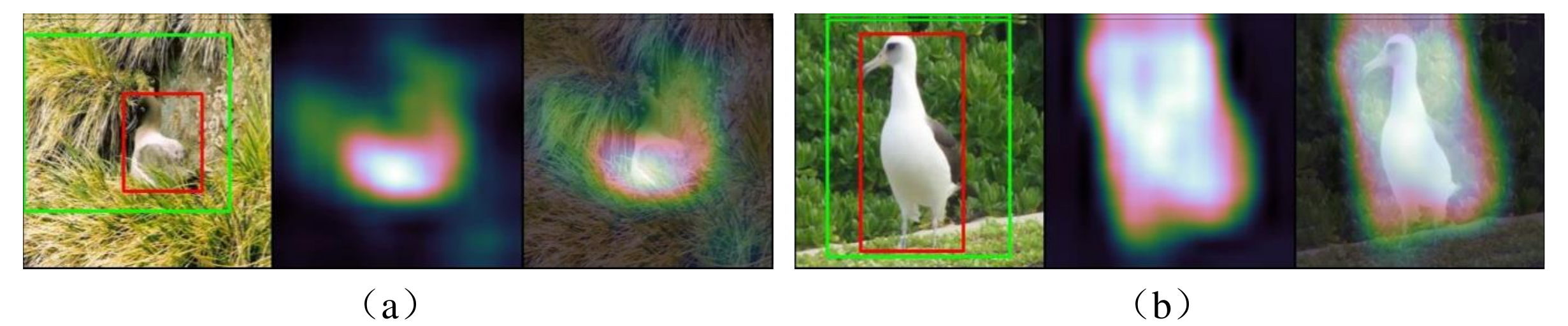}
\caption{Example images obtained by ResNet50-ADL. From left to right in each sub figure: the input
image, the heatmap, and the overlap between the heatmap and the input image.
In input image, the ground-truth bounding boxes are marked in red and the predicted are in green. The erasing operation sometimes leads to the attention spreading into the background. Meanwhile, the bounding box is too large to precisely
locate the object.
}
\label{fig:overfit}
\end{figure}

\begin{figure*}[t]
\centering
\includegraphics[height=5.8cm]{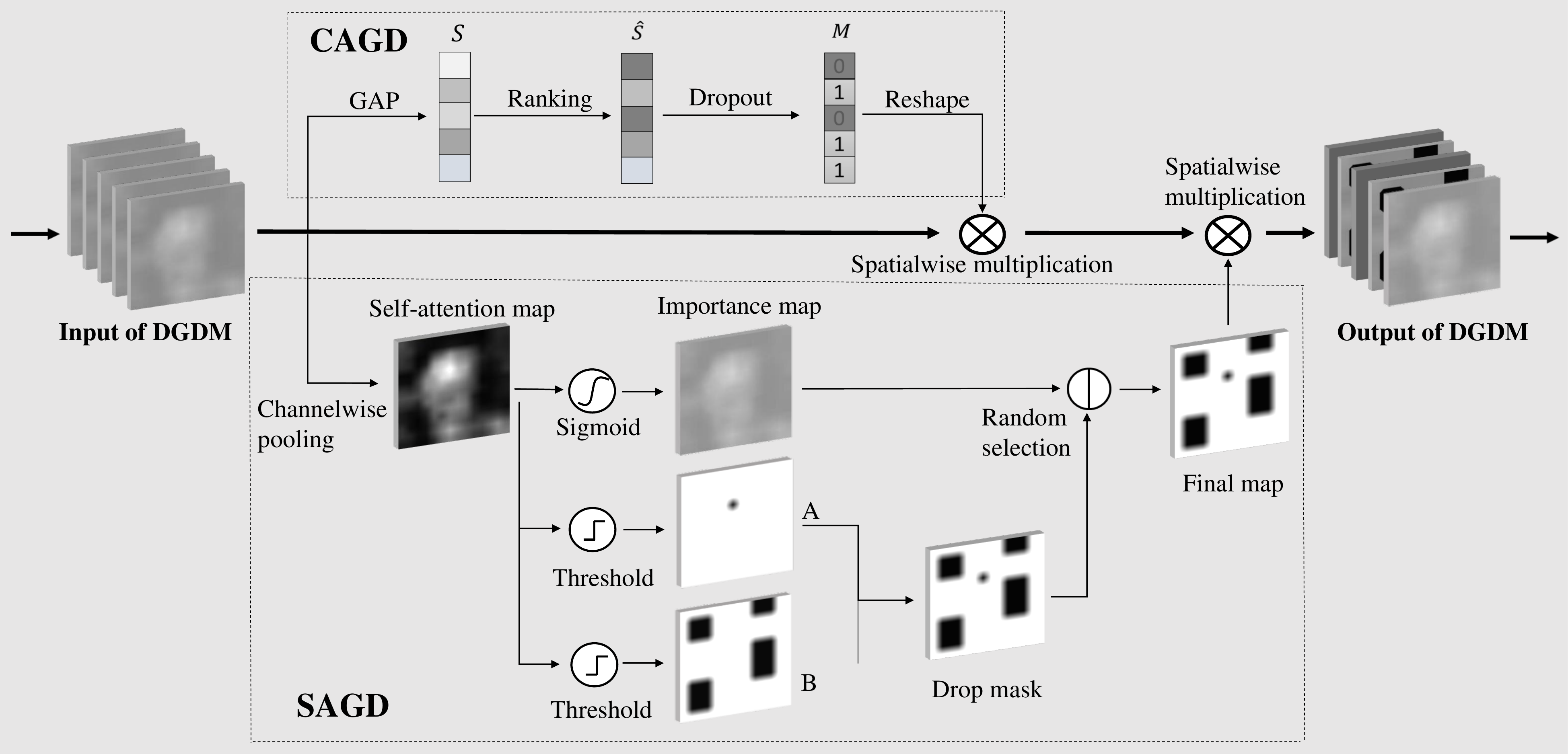}
\caption{Overall structure of the DGDM. It contains two key
components, CAGD and SAGD. In CAGD, we rank channel attention and consider the attentions with the top-k largest magnitudes as important ones.
Some low-valued elements are kept to increase their value if they become
important during training. For SAGD, the drop mask can not only efficiently erase the information by removing contiguous regions of feature maps rather than individual
pixels, but also sense the foreground objects and background regions to
alleviate the attention misdirection. The importance map is used to highlight the most discriminative regions of target object and suppress less useful ones. Finally, we
randomly select one of these two maps at each iteration and then multiply it to
the input feature map. It is worth noting that this figure shows the case when the drop mask is chose.
}
\label{fig:example}
\end{figure*}

Erasing the most discriminative parts is a simple yet poweful method
for WSOL. For example, ADL uses the
self-attention mechanism as supervision to encourage the model to learn the more useful information of the object. However, the erasing methods
abandon a lot of information on the most discriminative
regions. This forces the model to highlight the less discriminative
parts and sometimes captures useless information of the background,
which leads to the attention misdirection and the biased localization.
As shown in Figure \ref{fig:overfit}, the bounding box is too large to precisely locate the object, and the classification performance is not as good as before since the focused attention has been changed to other objects.

In this paper, we propose a dual-attention guided dropblock module (DGDM), a lightweight yet powerful method, for WSOL, which is illustrated in Figure~\ref{fig:example}.
It contains two key components, the channel attention guided dropout (CAGD) and the spatial attention guided dropblock (SAGD), to learn the discriminative and complementary
features by using the spatial and the channel attentions,
respectively. Specifically, in CAGD, we first compress the spatial information of input feature map by GAP to generate channel attention. We also rank
the obtained channel attention by a measure of importance (\emph{e.g.}, magnitude), and
then discard some elements with low importance. In addition, some low-valued elements are also kept to increase their
value if they become important during training.
For SAGD,
we perform channelwise average pooling on feature map to produce a self-attention map.
By thresholding the obtained attention map, we generate a drop
mask. It can not only efficiently erase the information by removing the contiguous regions of feature maps rather than the individual pixels, but also sense the foreground objects and the background regions under supervision with the confidence of different regions to alleviate the attention misdirection. In addition, an importance map is also generated by using a sigmoid activation on the attention map to highlight the most discriminative regions of the target object and suppress the less useful ones. We randomly select one of the two maps at each iteration and then multiply it to
the input feature map.

Deep networks implemented with DGDM incorporate the image classification and WSOL. In an end-to-end learning manner, the proposed method captures the complementary
and discriminative visual features for precise object
localization and achieves good result of
image classification.

The main contributions of the proposed method are:

(1) We propose a lightweight and efficient attention module (DGDM) that can be
easily employed to any layer of the CNNs
to achieve the good performance of WSOL.
The CAGD is proposed to model channel interdependencies. We rank channel attention and consider the attentions with
the top-$k$ largest magnitude as important ones. We also keep some low-valued elements to increase their value if they become important
during training.

(2) The SAGD is designed to generate a drop mask and an importance map. Importantly, this drop mask can efficiently erase the information by removing the
contiguous regions of the feature maps and sense the foreground objects and background regions for alleviating the attention misdirection.

(3) The proposed
approach can be employed to different CNNs classifiers and achieve
the state-of-the-art performance on several commonly used datasets.

\section{Related Work}

\subsubsection{Attention mechanism} Attention mechanism is a data processing method learnt from human perception process \cite{mnih2014recurrent}. It
does not process all the data in equal, but focuses more weights on the most informative parts \cite{mnih2014recurrent}. Attention mechanisms have demonstrated their utility across various fields, such as scene segmentation \cite{fu2019dual}, image localization and understanding \cite{jaderberg2015spatial,li2018tell}, fine-grained visual classification \cite{chang2020devil,li2020oslnet}, and image inpainting \cite{yu2018generative}. In particular, the self-attention mechanism \cite{vaswani2017attention} was firstly proposed to draw global dependencies of inputs and applied it in machine
translation.
Residual attention networks (RAN) \cite{wang2017residual} can generate attention-aware features by
adopting mixed attention modules with heavy parameters.
The squeeze-and-excitation module \cite{hu2018squeeze} was introduced to exploit the channel-interdependencies. The module
can use less parameter to extract attention, and allow the network to perform feature recalibration.
%through which it can learn to use global information to
%selectively emphasise informative features and suppress less
%useful ones.
Convolutional block attention module \cite{woo2018cbam} was proposed to emphasize the meaningful features by fusing the cross-channel and spatial information together.
However, these techniques require extra training parameters for obtaining the
attention map.

%Different from previous works, we extend the attention
%mechanism in the task of WSOL, and
%carefully design two types of attention modules to capture informative features for better feature representations.
%with intra-class compactness
% Comprehensive empirical
%results verify the effectiveness of our proposed method.

\subsubsection{Dropout in convolutional neural networks} Dropout \cite{hinton2012improving} has been proven to be a practical technique to alleviate overfitting in fully-connected neural networks, which drops neuron activations with
some fixed probability during training. All activations are
used during the test phase, but the output is scaled according to the dropout probability.
%This technique works well as an exponential
%number of smaller sub-networks and as a result discourages the co-adaptation of feature detectors.
%In the years since, many methods inspired by the original dropout technique have been proposed.
Many methods inspired by the original
dropout have been proposed. These methods include dropconnect \cite{wan2013regularization} and Monte Carlo dropout \cite{gal2016dropout} and many others.
However, regular dropout is less effective for convolutional layers.
This can largely be attributed to two factors.
The first is that convolutional layers require less
regularization since these layers have much less parameters
than fully-connected layers. The second factor is that
there is strong correlation between the spatially adjacent pixels on the convolutional layers and these neighbouring pixels have the same information. Hence, the pixel-based
dropout tends to abandon some information in the input, but the information can be passed on from the other adjacent pixels that are still active.% ADL \cite{choe2019attention}

In an attempt to apply a structured form of dropout to the convolutional layer, Cutout \cite{devries2017improved} drops out contiguous
regions of input images instead of individual pixels in the input
layer of CNNs. This method induces the network to better
utilize the contextual
information of the image, rather than relying
on a small set of specific features. Dropblock \cite{ghiasi2018dropblock} generalizes Cutout by applying Cutout at every feature map in
convolutional networks. Its main difference from regular dropout is that it discards
the contiguous regions from feature map rather than independent random units. ADL \cite{choe2019attention} utilizes attention mechanism
to find the maximally activated part and then drops them. However, the method only drops the strong activated pixels rather than strong activated region. %This will be further discussed in Subsection 3.3.

\subsubsection{Weakly supervised object localization} WSOL is an alternative cheaper way to identify the object location in a given image by solely using the image-level
supervision, \emph{i.e.}, presence
or absence of object categories \cite{cinbis2016weakly,zhou2016learning,bency2016weakly}. A WSOL method decomposes an image into a ``bag'' of
region proposals and iteratively selecting an instance (a proposal) from each
bag (an image with multiple proposals) to minimize the image
classification error in
step-wised manner \cite{cinbis2016weakly}. Recent research \cite{zhou2016learning} utilizes CNNs classifier for specifying the spatial distribution of discriminative patterns for different
image classes. A way to pursue the full object extent is self-paced
learning \cite{zhang2018self}. The self-produced guidance
(SPG) approach utilizes the CNNs to incorporate the
high confident regions, and the attention maps are then leveraged to
learn the object extent under the auxiliary supervision
masks of foreground and background regions. The other way to enhance the object localization is adversarial erasing \cite{zhang2018adversarial,singh2017hide,choe2019attention},
which first highlights the most discriminative regions on the feature map or input image and
then drops them so that the less discriminative regions can be
highlighted during training phase. Nevertheless,
most existing approaches use
alternative optimization, or requiring a lot of computing resources to erase the
most discriminative regions exactly.

\section{Dual-attention guided dropblock module }

\subsection{Spatial attention module}

Let $\bm{F}\in \mathbb{R}^{H\times W \times C}$ be a convolutional feature
map. Note that $C$ denotes the channel number, $H$ and $W$ are the height and the width of the feature map, respectively. For simplicity, the mini-batch dimension is omitted in this notation. We perform channelwise average pooling (CAP) on the input map $F$ to produce the 2D spatial attention map $\bm{M}_{self}\in \mathbb{R}^{H\times W}$.
$\bm{M}_{self}$ is then fed into a sigmoid function
to generate our importance map $\bm{M}_{imp}\in \mathbb{R}^{H\times W}$. The spatial attention focuses on where a discriminative part is. In short, the importance map is computed
as
\begin{align}
\bm{M}_{self} = CAP(\bm{F}),
\end{align}
\begin{align}
\bm{M}_{imp} = \sigma(\bm{M}_{self}),
\end{align}
where $\sigma$ denotes the sigmoid function.

Convolutional layers in the model are encouraged to generate
meaningful attention map for improving the classification accuracy. Therefore, in this attention map, the discriminative
power of model is proportional to the intensity of each pixel.
To make use of the information obtained in the CAP
operation, we follow it with the second operation by using the sigmoid function, and then
apply it to feature map.
In this way, the spatial distribution
of the most discriminative region can be approximated by this attention map efficiently, which improves the feature representation
for WSOL. We observe that the importance map $M_{imp}$ usually highlight the
most discriminative regions of target object and  suppress less useful ones.
In particular,
the most discriminative regions are some pixels whose intensity
is close to one, while the parts with low values are considered as the background.
Also, extra parameters are not required for our method to obtain the importance map.

\begin{figure}[t]
\centering
\includegraphics[height=2.5cm]{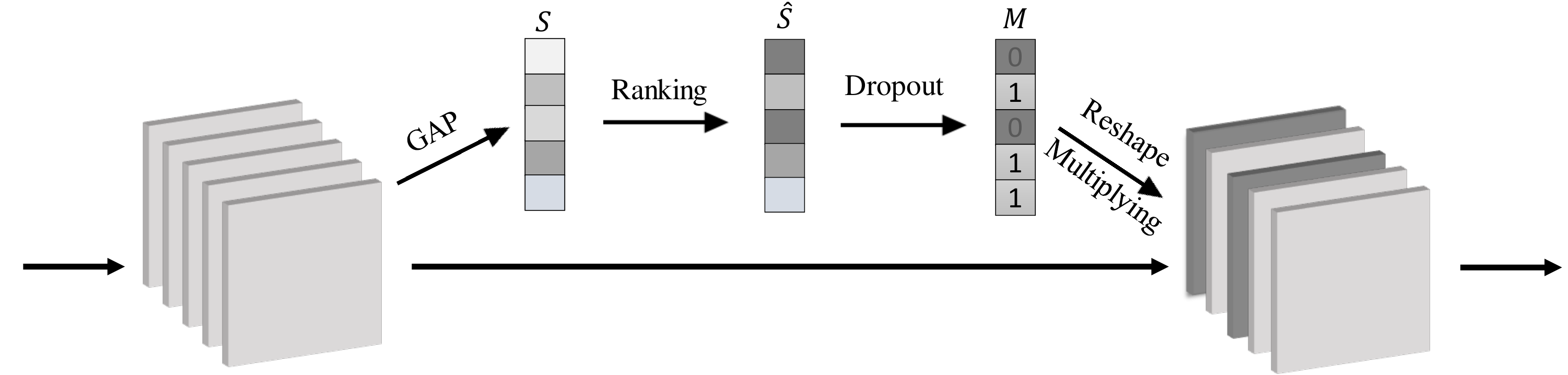}
\caption{Diagram of CAGD ($C=5$). As illustrated, we first compress spatial information of a feature map by GAP to generate
channel attention. We also rank
channel attention according to a fast measure of importance (magnitude), and then discard some elements with low importance. Regarding to
the channel selection, the binary mask is generated to indicate whether each channel is selected or not.}
\label{fig:channelblock}
\end{figure}

\subsection{Channel attention guided dropout}

As illustrated in Figure \ref{fig:channelblock}, we first gather the spatial information of a feature map $\bm{F}$ by performing global average pooling (GAP)
operation to generate the global information embedding: $\bm{S}\in\mathbb{R}^{C}$. As mentioned before, the discriminative power of the model is proportional to the intensity of each
pixel. Hence, the embedding can be considered as channel attention.
According to the relative magnitude of the attention, a binary mask is generated to indicate whether each channel is selected or not. This attention-guided pruning strategies can be treated as a special way to model
the interdependencies across the channels. We rank the
channel attentions by a fast and approximate measure of importance (magnitude), and
then discard those elements with low importance. The strategy considers
the attentions with
the top-$k$ largest magnitude as important ones.
We also treat each element of $\bm{S}$ separately under the $L_{1}$-norm.
\begin{align}
\widehat{\bm{S}} =\{ \arg \max_k |S_{c}|, S_{c}\in \bm{S} \},
\end{align}
where $\arg \max_{k}$ is used to return
the top-$k$ elements out of all elements being considered.

Inspired by Target Dropout \cite{gomez2018targeted}, we keep only the $k$ channels of highest magnitude in the
network and drop out the others. Similar to regular dropout, this encourages the network to learn a sparse representation. However, we allow some low-valued elements to increase their value during training. As in Target Dropout \cite{gomez2018targeted}, we introduce the stochasticity into the process using a drop probability $\alpha$ and a drop threshold $\beta$, which is the prefixed multiple of minimum value of channel attentions. Then, we obtain the drop mask $\bm{M}_{drop1}\in \mathbb{R}^{C}$
by setting each element as 0 with probability $\alpha$ if it is smaller than the drop threshold $\beta$,
and 1 if it is larger. To address the difficulty of learning additional
parameters, we set $\alpha$ to 0.5 in our work.

\subsection{Spatial attention guided dropblock}

%The importance
%map $\bm{M}_{imp}$ obtained by the spail attention module rewards the most discriminative region for increasing the
%classification power of the model. Unfortunately, the classifiers
%tend to focus only on the most discriminative features
%to increase their classification accuracy. Previous
%study \cite{singh2017hide} has reported that losing some classification accuracy results in the huge boost in localization performance. This is caused by the usage
%of a drop mask, which erases the most
%discriminative region of the object. From this idea, ADL \cite{choe2019attention} produces a self-attention map from input feature map,
%and generates a drop mask and an importance map for inducing the model to capture
%the integral extent of the object. During training phase, the importance map or drop
%mask is stochastically selected. ADL has two
%main parameters: a drop threshold $\gamma$ and drop-rate. The $\gamma$
%controls the size of the region to be dropped, and the drop rate indicates
%how frequently the drop mask is applied.
We observe that ADL drops only strongly activated parts according to the drop threshold. In fact, similar to the pixel-based dropout, the ADL is not really the region-based dropout. This is because the drop mask of ADL is generated by setting each pixel as 1 if it is smaller than the drop threshold $\gamma$,
and 0 if it is larger. This means that $\gamma$ determines how many
activation units should be discarded. However,
there is the strong correlation between the neighbouring pixels on the convolutional layers, and these adjacent pixels have the similar
information. Hence, the ADL cannot entirely remove the information.

\begin{figure}[t]
\centering
\includegraphics[height=4cm]{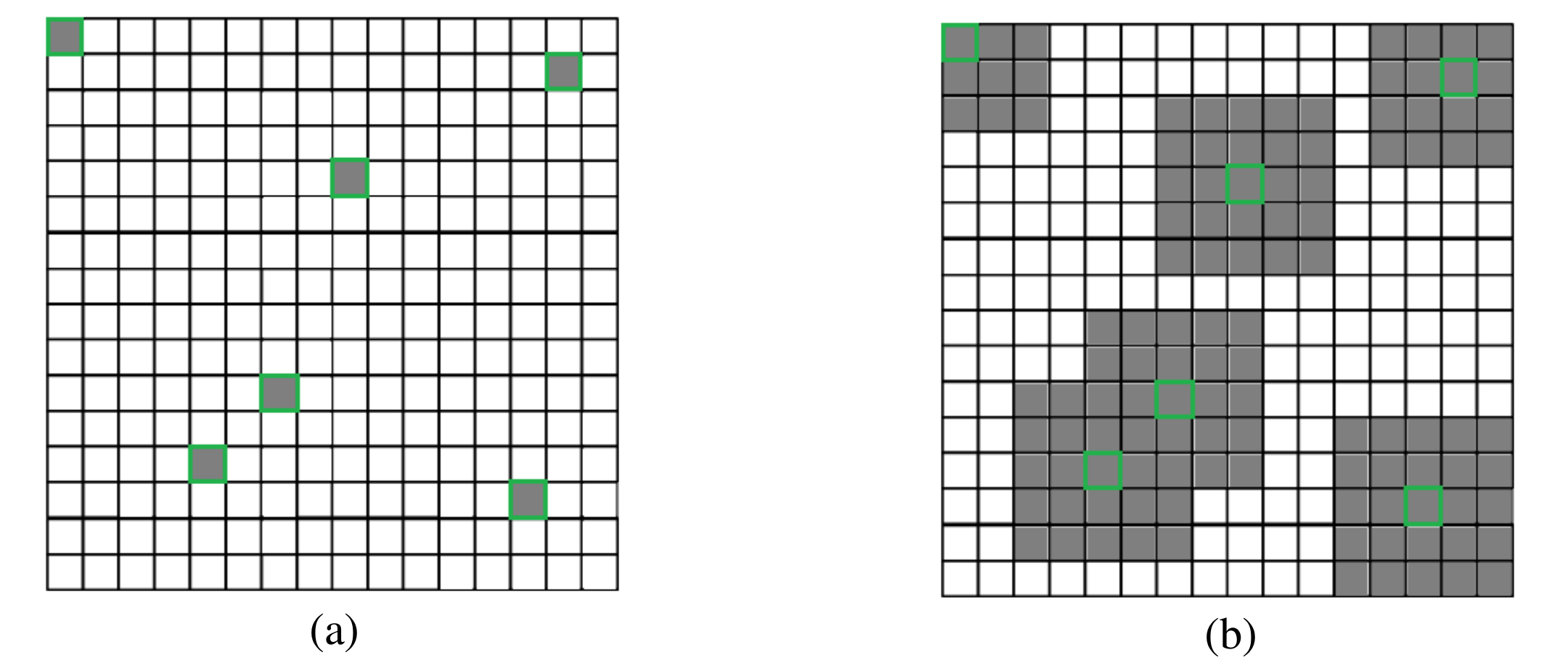}
\caption{(a) Original drop mask; (b) our drop mask.}
\label{fig:block}
\end{figure}

To address above problems, we propose a region-based dropout in a similar fashion
to regular dropout, but with an important distinction. The difference is that we drop the contiguous regions of feature maps rather than the individual pixels, as illustrated in Figure \ref{fig:block}.
Its main difference from Dropblock \cite{ghiasi2018dropblock} is that its drop mask $\bm{M}_{drop2}$ is calculated from $\bm{M}_{self}$ obtained by the spatial attention module.
Then,
the shared drop mask across different feature channels or each feature
channel has the same drop mask.
The proposed method has two main hyperparameters: $block\_size$ and $\delta$. The $block\_size$ presents the size of the block to be discarded, and $\delta$ determines how many
activation units to discard. When $block\_size$ equals to 1 and covers the full feature map, the region-based dropout reduces to the standard dropout
and SpatialDropout \cite{tompson2015efficient}, respectively.
This technique can efficiently
remove the information on the feature map. Hence, it forces the network to better
capture the full context of the feature map, rather than relying
on the presence of a small set of the discriminative features.

It is well known that we can divide images into background and foreground regions.
Also, the object of interests is usually consisted of the foreground pixels.
The work \cite{zhang2018self} has reported that the attention map stands for the probabilities of corresponding pixel to be background or foreground. The initial background
and object can be produced through the values in the self-attention maps. In particular, the regions with very large values are foreground, while the regions with small values are considered as background.
Removing discriminative regions forces the CNNs model to capture the less discriminative
part, which sometimes leads to the attention misdirection and the
biased localization. Base on this, we can sense background parts and the foreground objects by using the drop mask $\bm{M}_{drop2}$ according to the self-attention map, which will finally benefit WSOL.

We define $\bm{M}_{drop2}$ as follows. We set $\bm{M}_{drop2}^{x,y}=0$ if the
pixel at $x_{th}$ row and $y_{th}$ column belongs to background parts or the most discriminative parts. %$\bm{M}_{drop2}^{x,y}=1$ if it belongs to object regions.
Otherwise we have $\bm{M}_{drop2}^{x,y}=1$.
Specifically, $\bm{M}_{drop2}$ can be calculated by
\begin{align}
\bm{M}_{drop2}^{x,y} =
\begin{cases}
0,& \bm{M}_{self}^{x,y} < \delta_{l}, \bm{M}_{self}^{x,y} > \delta_{h} ,\\
1,& \text{otherwise,}
\end{cases}
\end{align}
where $\delta_{l}$ and $\delta_{h}$ are thresholds to identify the regions in feature maps as the background and the most discriminative parts of foreground, respectively.

\subsection{Network Implementation}

Fully CNNs with DGDM fuse the complementary
discriminative parts for precise object localization
and the accurate image classification in an end-to-end manner. Following the work in \cite{choe2019attention}, the DGDM is inserted in higher-level feature maps of the CNNs model.

For the DGDM, convolutional feature maps are averaged by channelwise average pooling to generate the meaningful self-attention map. The attention maps are then activated by using a sigmoid function.
We then
multiply the obtained importance map to input.
Based on this attention map, the CNNs model can distinguish the foreground objects from background regions by multiplying $\bm{M}_{drop2}$ to the input feature map.
The importance map $\bm{M}_{imp}$ highlights the most discriminative information for achieving the
good accuracy of the model. Unfortunately, the classifiers
often rely only on the most discriminative information. The work in \cite{singh2017hide} has reported that losing some classification accuracy results in the huge boost in localization performance. This is owing to the application of a drop mask, which erases the most
discriminative parts. From this idea, the importance map or drop
mask is stochastically selected during training phase. As presented in ADL \cite{choe2019attention}, we also introduce a drop rate, which controls how frequently the drop mask is employed. In addition,
channel attention guided mask $\bm{M}_{drop1}$ is also applied to input feature maps to model the interdependencies across the channels.

The CNNs model is optimized with Stochastic
Gradient Descent (SGD) algorithm.
We extract the heatmap using the same method as introduced in \cite{zhou2016learning}.
Finally, a thresholding
method [38] is then directly applied to predict the locations of target object.

\section{Experiment}
\label{sec:blind}

\subsection{Experimental Setup}

\subsubsection{Datasets} We evaluate the performance of the proposed
method in the commonly used
CUB-200-2011 \cite{wah2011caltech}, Stanford Cars \cite{krause20133d}, and ILSVRC 2016 \cite{russakovsky2015imagenet} datasets.
%The CUB-200-2011 contains 200 species of birds
%with 5,994 images for training and 5,794 for testing.
%The Stanford Cars dataset \cite{krause20133d} includes
%16,185 images with 196 categories of cars.
%This data is split into
%8,144 training images and 8,041 testing images.
%The ILSVRC is a dataset, which includes
%1,000 different categories.
%We train the model with 1.3 million images
%for training, and 50,000 images in the validation
%set for testing.

%Some
%example images are shown in Figure \ref{fig:data}.
%Input size
%for the three datasets is set to $224\times224$.

%\begin{figure}[t]
%\centering
%\includegraphics[height=3.5cm]{data}
%\caption{Example images.(a)CUB-200-2011. (b)Stanford
%Cars. (c)ImageNet-1k.}
%\label{fig:data}
%\end{figure}

\subsubsection{Evaluation metrics} Three evaluation metrics are used for WSOL
evaluation \cite{singh2017hide}. They are localization accuracy
with known ground-truth class
(GT-Loc), Top-1
classification accuracy (Top-1 Clas), and Top-1
localization accuracy (Top-1 Loc).
%Top-1 classification accuracy (Top-1 Clas) determines
%that the answer is correct when the estimated class is equal
%to the ground truth class.
%measure the impact of the proposed method on image classification performance.
%Localization accuracy
%with known ground-truth class
%(GT-known Loc) considers
%the answer as correct when the intersection over union (IoU) between
%the ground truth bounding box and predicted box for the
%ground truth class is $50\%$ or more. Top-1
%localization accuracy (Top-1 Loc) determines
%that the answer is correct when both GT-known Loc and Top-1 Clas are correct.
It is worth noting that the most appropriate metric is Top-1 Loc for evaluating the performance of WSOL.

\subsubsection{Experimental details} The proposed DGDM is
integrated with the commonly used CNNs including VGG \cite{simonyan2014very}, ResNet \cite{he2016deep},
ResNet-SE \cite{hu2018squeeze}, and InceptionV3 \cite{szegedy2016rethinking}. Following the settings of work \cite{choe2019attention}, the drop rate is set as 75\%, and apply DGDM to higher-level and intermediate layers
of CNNs. These networks are initialized with ImageNet pre-trained weights.

\subsection{Ablation studies}

The ablation studies on CUB-200-2011 with the pre-trained VGG-GAP are used to evaluate the effects of the proposed DGDM.
Furthermore, we conduct some experiments on Stanford Cars to investigate the effect of network depth on localization performance.
During training phase, the DGDM is inserted in all the pooling layers and the conv5-3 layer.

% Table generated by Excel2LaTeX from sheet 'Sheet1'
\begin{table*}[t]
  \centering
  \caption{Upper and Middle: Accuracy according to different $block\_size$. Lower: Accuracy according to different drop threshold $\beta$. $\bm{M}_{drop2}$\_stage1: erasing the most discriminative
parts of the object. $\bm{M}_{drop2}$\_stage1+$\bm{M}_{drop2}$\_stage2: removing the most discriminative
parts and alleviating attention misdirection. $\bm{M}_{drop2}$\_stage1+$\bm{M}_{drop2}$\_stage2+$\bm{M}_{drop1}$:
removing the most discriminative
parts of the object, alleviating attention misdirection, and utilizing channel attention guided dropout.
Adap\_7: adaptive and $block\_size$ calculated by [H(W)
$/7$].
}
    \begin{tabular}{lcccc}
    \toprule
     Method     & \multicolumn{1}{l}{$block\_size$} & \multicolumn{1}{l}{GT } & \multicolumn{1}{l}{Top-1 } & \multicolumn{1}{l}{Top-1} \\
          &       & \multicolumn{1}{l}{Acc(\%)} & \multicolumn{1}{l}{Clas(\%)} & \multicolumn{1}{l}{ Loc  (\%)} \\
\cmidrule{2-5}     & 1     & 62.70    & 71.99 & 45.20  \\
\multicolumn{1}{l}{$\bm{M}_{drop2}$\_stage1}          & 2     & 73.48  & 68.30 & \textbf{52.57}\\
          & 3     & 69.53   & 55.56  & 43.10  \\
          & Adap\_7 & \textbf{73.52}  & 65.83 & 50.46  \\
    \midrule
          & 1     & 73.48  & 69.11  & 52.11  \\
          & 2     & 73.94  & 69.68  & 53.27   \\
    \multicolumn{1}{l}{$\bm{M}_{drop2}$\_stage1+$\bm{M}_{drop2}$\_stage2} & 3     & \textbf{74.23}  & 69.00 & \textbf{53.79}  \\
          & 4     & 73.48  & 69.11  & 52.11 \\
          & Adap\_7 & 72.02 & 	69.21  &51.42 \\
    \midrule
          & $\beta$ & \multicolumn{1}{l}{GT } & \multicolumn{1}{l}{Top-1 } & \multicolumn{1}{l}{Top-1} \\
          &       & \multicolumn{1}{l}{Acc(\%)} & \multicolumn{1}{l}{ Clas (\%)} & \multicolumn{1}{l}{ Loc (\%)} \\
\cmidrule{2-5}     & 2     & 73.44  & 69.19  & 52.56 \\
\multicolumn{1}{l}{$\bm{M}_{drop2}$\_stage1+$\bm{M}_{drop2}$\_stage2+$\bm{M}_{drop1}$}          & 2.5   & 74.88  & 69.50 & 54.31 \\
          & 3     & \textbf{75.02}  & 69.89 & \textbf{54.34} \\
          & 3.5   & 72.65  & 69.68 & 52.67 \\
    \bottomrule
    \end{tabular}%
  \label{tab:ablation}%
\end{table*}%

The drop mask map $\bm{M}_{drop2}$ can not only remove a small set of the discriminative features to better capture the full context of the feature map, but also identify the regions in the feature maps as the background and the foreground, respectively. First, we verify the effectiveness of removing importance parts on accuracy. The upper block of Table \ref{tab:ablation} presents the experimental results when different $block\_size$ are adopted.
From these results, it can be seen that we achieve the best localization accuracy
when $block\_size$ is $2$. We also present the result when $block\_size$ is adaptive and calculated by [H(W)
$/7$]. It can be observed that the drop
masks with $block\_size$ = 2 remove the discriminative
region more precisely than those with other $block\_size$.
We observe that the classification accuracy decreases as $block\_size$ increases. This is because the model never captures
the most discriminative region.
%Classification accuracy decreases significantly, which adversely
%leads to the huge boost in
%localization performance.

Next, we investigate the effect of removing a small set of background on accuracy. The middle block of Table \ref{tab:ablation} presents the results when we use different $block\_size$. It can be observed that the best object localization accuracy can be established
when $block\_size$ is $3$. In addition, the Top-1 Loc increases again (from 52.57\% to 53.79\%) and the classification accuracy achieve
$0.7\%$ improvement (from 68.30\% to 69.00\%). This indicates that removing a small set of background boosts the performance
of WSOL. The reason lies in that the proposed erasing method does not lead to the attention misdirection when the discriminative parts are erased.

Thirdly, we observe the effect of channel attention guided dropout on the
accuracy by using four different drop thresholds. The lower block of Table \ref{tab:ablation} reports experimental
results. Based on this, we can conclude that the value of the drop threshold has an important effect on performance of WSOL.
It can also be seen that three evaluation metrics are improved when the drop threshold is 3.

\begin{table}[t]
  \centering
  \caption{Performance comparison on Stanford Cars test set with several popular baseline architectures.}
    \begin{tabular}{cccc}
    \toprule
    \toprule
 Method  & Backbone & \multicolumn{1}{l}{Top-1 Loc} & \multicolumn{1}{l}{Top-1 Clas} \\
    \midrule
 ADL  & ResNet18 & 86.50 & 88.50  \\
 Ours  & ResNet18 & \textbf{87.38}& 88.91 \\
    \midrule
ADL   & ResNet34 & 87.37 & 90.24 \\
Ours   & ResNet34 & \textbf{88.44} & 89.91 \\
    \midrule
 ADL   & ResNet50&  89.32& 91.48 \\

Ours   & ResNet50& \textbf{90.25} & 91.85 \\

    \bottomrule
    \bottomrule
    \end{tabular}%
  \label{tab:Stanford Cars}%
\end{table}%

Lastly, we generalise our method to other dataset (Stanford Cars) and
investigate the effect of network depth on localization accuracy.
Some experiments are performed with several popular baseline architectures (ResNet-18, ResNet-34, ResNet-50).
DGDM is integrated into these networks.
This method and its counterpart (ADL) are
trained on the Stanford Cars dataset.
We report the performance
of our method and its counterpart on Stanford Cars
in Table \ref{tab:Stanford Cars}. It can be
observed that in all the comparisons the proposed method outperforms its counterpart and the proposed method obtains the best performance when ResNet-50 is employed as backbone. This suggests that the benefits of DGDM are not limited to network depth and certain dataset.

\subsection{Comparison with the state-of-the-arts}

We compare our proposed approach with existing WSOL techniques on the CUB-200-2011 test set and ILSVRC validation set and give the results in Table \ref{tab:CUB-200-2011}, Table \ref{tab:Imagenet-1k}, and Table \ref{tab:Stanford Cars}, respectively. These approaches include CAM
\cite{zhou2016learning}, DANet \cite{xue2019danet}, ACoL \cite{zhang2018adversarial}, SPG \cite{zhang2018self}, and ADL \cite{choe2019attention}.

% Table generated by Excel2LaTeX from sheet 'Sheet1'
\begin{table*}[htbp]
  \centering
  \caption{Quantitative evaluation results on CUB-200-2011 test set with the state-of-
the-art results.}
    \begin{tabular}{cccccccc}
    \toprule
    Method & Backbone & FLOPs & \# of & \multicolumn{2}{c}{Overheads} & \multicolumn{2}{c}{CUB-200-2011} \\
\cmidrule{5-8}       &   &  (Gb)     & Params & Computation(\%) & Parameter(\%) & \multicolumn{1}{l}{Top-1 } & \multicolumn{1}{l}{Top-1}  \\
          &     &   & (Mb)  &       &       & \multicolumn{1}{l}{Loc  (\%)} & \multicolumn{1}{l}{ Clas (\%)}  \\
    \midrule
    CAM   & VGG-GAP &18.20 & 29.08    & 0     & 0     & 34.41 & 67.55  \\
    ACoL  & VGG-GAP &31.98 & 37.63   & 71.51 & 75.71 & 45.92 & 71.90 \\
    ADL   & VGG-GAP &18.20 & 29.08    & 0     & 0     & 52.36 & 65.27 \\
    DANet & VGG-GAP &24.12 & 48.56  & 32.53 &  66.99 & 52.52 & 75.40   \\
    Ours  & VGG-GAP &18.20 & 29.08    & 0     & 0     & \textbf{54.34} & 69.85  \\
    \midrule
%    Baseline & ResNet50 & 100   & 0     & 0     & 51.20  & 81.24 \\
    ADL   & ResNet50 &62.32 & 23.92   & 0     & 0     & 46.29 & 79.72  \\
    DANet   &ResNet50&74.33 & 32.63   & 19.27     & 36.41     & 51.10 & 81.60  \\
    Ours  & ResNet50 &62.32 & 23.92   & 0     & 0     & \textbf{59.40} & 76.20 \\
    \midrule
    CAM   & InceptionV3 &4.84 & 25.69   & 0     & 0     & 43.67 &    -    \\
    SPG   & InceptionV3 &31.98 & 37.63   &560.74 & 46.48  & 46.64 &    -   \\
    ADL   & InceptionV3 &4.84 & 25.69   & 0     & 0     & \textbf{53.04} & 74.55   \\
    DANet & InceptionV3 &7.23 & 30.62  & 49.38 &18.47  & 49.45 & 71.20  \\
    Ours  & InceptionV3 &4.84 & 25.69   & 0     & 0     & 52.62 & 72.23  \\
    \bottomrule
    \end{tabular}%
  \label{tab:CUB-200-2011}%
\end{table*}%

\subsubsection{CUB-200-2011} Table \ref{tab:CUB-200-2011} summaries the quantitative evaluation results on the CUB-200-2011 test set. With a VGG-GAP backbone,
our method reports $4.58\%$ higher Top-1 Clas and $1.98\%$ higher Top-1 Loc compared with the ADL approach \cite{choe2019attention}. With a ResNet50
backbone, it reports 8.30\% performance gain over
the DANet approach \cite{xue2019danet} at the cost
of little classification performance. This is the outcome of the trade-off relationship
between classification and localization accuracy
discussed in Subsection 3.4.
Our method significantly outperforms all the existing approaches, and obtains a new SOTA localization accuracy (59.40\%) even when other two backbones are
used. When InceptionV3 is used
as a backbone, this method still has comparable accuracy.

In addition to obtaining a better performance of WSOL, this method has high efficiency. Table \ref{tab:CUB-200-2011} also presents the number of parameters
as well as parameter and computation overheads. Similar to ADL \cite{choe2019attention}, additional training parameters are not required for our method,
and there are no computation overheads upon the backbone network.

\begin{figure*}[t]
\centering
\includegraphics[height=6.5cm]{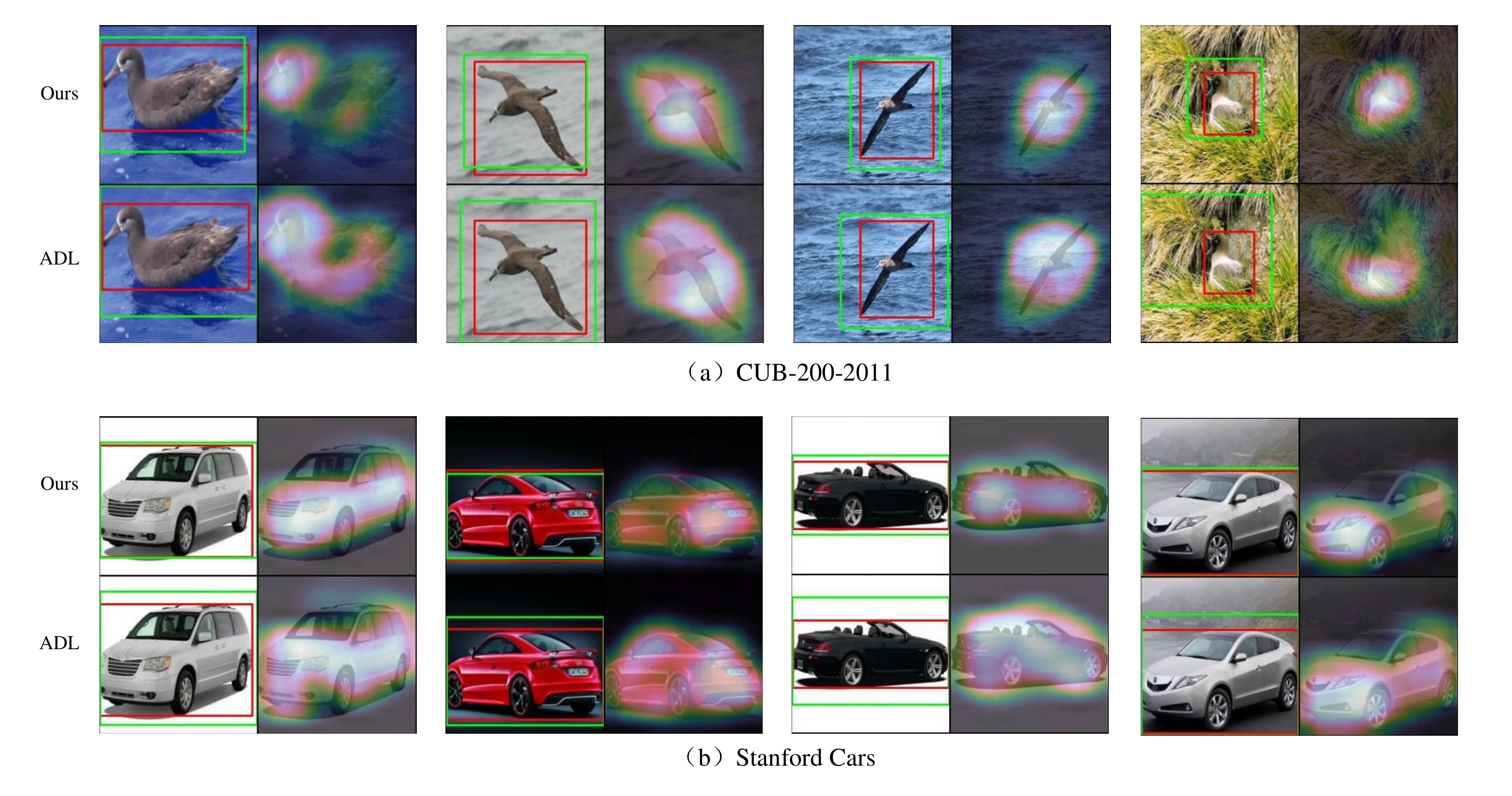}
\caption{Visualization results of ResNet50 on CUB-200-2011 and Stanford Cars.
We compare the visualization
results of DGDM-integrated network (ResNet50 + DGDM) with ADL.
From left image and right image in each figure: input
image and the overlap between the heatmap and the input image.
In input image, the ground-truth bounding boxes are in red and the predicted are in green.
%The left image in each figure is input
%image. The ground-truth bounding boxes are in red and the predicted are in green.
%The middle image is heatmap and the right image
%presents the overlap between the heatmap and the input image.
}
\label{fig:Visualization}
\end{figure*}

%\begin{figure*}[t]
%\centering
%\includegraphics[height=4cm]{mask}
%\caption{The self-attention map and the drop mask at higher-level layer of ResNet50. Insertion position of DGDM is named
%by the following scheme: stageID\_blockID.}
%\label{fig:Drop mask and self-attention map}
%\end{figure*}

% Table generated by Excel2LaTeX from sheet 'Sheet2'
\begin{table}[t]
  \centering
  \caption{Quantitative evaluation results on ILSVRC validation set with the state-of-the-art results.}
    \begin{tabular}{cccc}
    \toprule
    \toprule
Method  & Backbone  & \multicolumn{1}{l}{Top-1 Loc} & \multicolumn{1}{l}{Top-1 Clas} \\
    \midrule
% Backprop &  VGGNet &38.88 & -  \\
%CAM  &  VGGNet & 42.80  & 66.60 \\
%ACoL  &  VGGNet & \textbf{45.83} & 67.50 \\
%ADL  &  VGGNet & 44.92 & 69.48 \\
%Ours  & VGGNet &   42.77    & 65.74 \\
%    \midrule
CAM  &  ResNet50-SE & 46.19 & 76.56 \\
ADL  &  ResNet50-SE & 48.53 & 75.56 \\
Ours &ResNet50-SE & \textbf{48.81} & 73.50 \\
%    \midrule
%    InceptionV3-CAM & 46.29 & - \\
%    InceptionV3-SPG & 48.60 & 75.56 \\
%    InceptionV3-ADL  & 48.71 & 72.83 \\
%    InceptionV3-DANet & 47.53 & 72.50 \\
%    Ours  &  &  \\
    \bottomrule
    \bottomrule
    \end{tabular}%
  \label{tab:Imagenet-1k}%
\end{table}%

\subsubsection{ILSVRC} In Table \ref{tab:Imagenet-1k}, we evaluate the performance of our proposed method on the large-scale ILSVRC dataset.
%By adopting the VGG-GAP backbone,
%our method outperforms Backprop significantly, but is slightly worse than other existing techniques. However,
With a ResNet50-SE backbone, our method obtains a better localization performance than ADL and CAM. To sum up, our method achieves new state-of-the-art accuracy compared with the current techniques.

\subsubsection{Discussion} The work in \cite{choe2019attention} has reported
that ADL extracts the discriminative
information of the background that often appears with the object. To investigate the substantial difference
between the proposed method and ADL, we
show the heatmaps as well as the predicted bounding boxes on Stanford Cars and CUB-200-2011 in Figure \ref{fig:Visualization}. It can be observed that the object
localization maps produced by our method can
obtain more accurate bounding boxes than ADL.
That is, the network implemented with DGDM learns well to exploit the information in the target object regions and learn patterns from them.
We attribute it to the removing operation which induces the network to discover more discriminative features to achieve better performance of WSOL.
On the CUB-200-2011 dataset,
%all the categories belong to birds. Therfore, similar backgrounds occur in spite of the classes (\emph{e.g.}, sea). This indicates that
the less discriminative part sometimes is the background frequently occurring with the object
\cite{choe2019attention}. For example, our method and ADL can discover nearly entire parts of a bird, \emph{e.g.}, the wing and head, but ADL also learns more background information when the most discriminative information
is discarded. %For example, ADL captures not only the bird, but also the sea.

%Figure \ref{fig:Drop mask and self-attention map} gives the self-attention map and the drop mask at higher-level layer of ResNet-50.
%The self-attention maps generated from the classification
%network are applied as supervision to distinguish the target objects and background regions.
%It can be observed that the attention regions spread to the background with ADL method while our method alleviate the attention misdirection.
%The proposed drop mask can hide the most discriminative part and remove a small set of background
% by erasing contiguous regions of feature maps rather than
%independent individual pixels. This
%guides the CNNs model to discover the less discriminative regions better,
%rather than focusing only on the most
%discriminative regions to improve classification accuracy.

\section{Conclusions}

In this paper, we proposed a simple yet effective dual-attention guided dropblock module (DGDM) for  weakly supervised object localization (WSOL). We designed two key components of DGDM, namely the channel attention guided dropout (CAGD) and the spatial attention guided dropblock (SAGD), and integrated them with the deep learning framework. The proposed method hides the most discriminative part and then encourages the CNNs model to discover the less discriminative part. We defined a pruning strategy so that CAGD can be adapted to model the interdependencies across the channels. In addition, SAGD can not only efficiently remove the information by erasing the contiguous regions of feature maps rather than the independent individual pixels, but also sense the target objects and background regions to alleviate the attention misdirection. Compared to some existing WSOL techniques, the proposed method is lightweight, and can be easily employed to different CNNs classifiers. We also have achieved new SOTA localization accuracy on CUB-200-2011, Stanford Cars, and ILSVRC.

%\clearpage
% ---- Bibliography ----
%
% BibTeX users should specify bibliography style 'splncs04'.
% References will then be sorted and formatted in the correct style.
%
\bibliographystyle{splncs04}
\bibliography{egbib}

\begin{thebibliography}{10}
\providecommand{\url}[1]{\texttt{#1}}
\providecommand{\urlprefix}{URL }
\providecommand{\doi}[1]{https://doi.org/#1}

\bibitem{bency2016weakly}
Bency, A.J., Kwon, H., Lee, H., Karthikeyan, S., Manjunath, B.: Weakly
  supervised localization using deep feature maps. In: Proceedings of the
  European Conference on Computer Vision (ECCV). pp. 714--731. Springer (2016)

\bibitem{chang2020devil}
Chang, D., Ding, Y., Xie, J., Bhunia, A.K., Li, X., Ma, Z., Wu, M., Guo, J.,
  Song, Y.Z.: The devil is in the channels: Mutual-channel loss for
  fine-grained image classification. IEEE Transactions on Image Processing
  \textbf{29},  4683--4695 (2020)

\bibitem{choe2019attention}
Choe, J., Shim, H.: Attention-based dropout layer for weakly supervised object
  localization. In: Proceedings of the IEEE Conference on Computer Vision and
  Pattern Recognition (CVPR). pp. 2219--2228 (2019)

\bibitem{cinbis2016weakly}
Cinbis, R.G., Verbeek, J., Schmid, C.: Weakly supervised object localization
  with multi-fold multiple instance learning. IEEE Transactions on Pattern
  Analysis and Machine Intelligence  \textbf{39}(1),  189--203 (2016)

\bibitem{devries2017improved}
DeVries, T., Taylor, G.W.: Improved regularization of convolutional neural
  networks with cutout. arXiv preprint arXiv:1708.04552  (2017)

\bibitem{fu2019dual}
Fu, J., Liu, J., Tian, H., Li, Y., Bao, Y., Fang, Z., Lu, H.: Dual attention
  network for scene segmentation. In: Proceedings of the IEEE Conference on
  Computer Vision and Pattern Recognition (CVPR). pp. 3146--3154 (2019)

\bibitem{gal2016dropout}
Gal, Y., Ghahramani, Z.: Dropout as a bayesian approximation: Representing
  model uncertainty in deep learning. In: International Conference on Machine
  Learning (ICML). pp. 1050--1059 (2016)

\bibitem{ghiasi2018dropblock}
Ghiasi, G., Lin, T.Y., Le, Q.V.: Dropblock: A regularization method for
  convolutional networks. In: Advances in Neural Information Processing Systems
  (NIPS). pp. 10727--10737 (2018)

\bibitem{gomez2018targeted}
Gomez, A.N., Zhang, I., Swersky, K., Gal, Y., Hinton, G.E.: Targeted dropout

\bibitem{he2016deep}
He, K., Zhang, X., Ren, S., Sun, J.: Deep residual learning for image
  recognition. In: Proceedings of the IEEE Conference on Computer Vision and
  Pattern Recognition (CVPR). pp. 770--778 (2016)

\bibitem{hinton2012improving}
Hinton, G.E., Srivastava, N., Krizhevsky, A., Sutskever, I., Salakhutdinov,
  R.R.: Improving neural networks by preventing co-adaptation of feature
  detectors. arXiv preprint arXiv:1207.0580  (2012)

\bibitem{hu2018squeeze}
Hu, J., Shen, L., Sun, G.: Squeeze-and-excitation networks. In: Proceedings of
  the IEEE Conference on Computer Vision and Pattern Recognition (CVPR). pp.
  7132--7141 (2018)

\bibitem{jaderberg2015spatial}
Jaderberg, M., Simonyan, K., Zisserman, A., et~al.: Spatial transformer
  networks. In: Advances in Neural Information Processing Systems (NIPS). pp.
  2017--2025 (2015)

\bibitem{krause20133d}
Krause, J., Stark, M., Deng, J., Fei-Fei, L.: 3d object representations for
  fine-grained categorization. In: Proceedings of the IEEE International
  Conference on Computer Vision Workshops. pp. 554--561 (2013)

\bibitem{li2018tell}
Li, K., Wu, Z., Peng, K.C., Ernst, J., Fu, Y.: Tell me where to look: Guided
  attention inference network. In: Proceedings of the IEEE Conference on
  Computer Vision and Pattern Recognition (CVPR). pp. 9215--9223 (2018)

\bibitem{li2020oslnet}
Li, X., Chang, D., Ma, Z., Tan, Z.H., Xue, J.H., Cao, J., Yu, J., Guo, J.:
  Oslnet: Deep small-sample classification with an orthogonal softmax layer.
  IEEE Transactions on Image Processing  (2020)

\bibitem{mnih2014recurrent}
Mnih, V., Heess, N., Graves, A., et~al.: Recurrent models of visual attention.
  In: Advances in Neural Information Processing Systems (NIPS). pp. 2204--2212
  (2014)

\bibitem{russakovsky2015imagenet}
Russakovsky, O., Deng, J., Su, H., Krause, J., Satheesh, S., Ma, S., Huang, Z.,
  Karpathy, A., Khosla, A., Bernstein, M., et~al.: Imagenet large scale visual
  recognition challenge. International Journal of Computer Vision
  \textbf{115}(3),  211--252 (2015)

\bibitem{simonyan2014very}
Simonyan, K., Zisserman, A.: Very deep convolutional networks for large-scale
  image recognition. arXiv preprint arXiv:1409.1556  (2014)

\bibitem{singh2017hide}
Singh, K.K., Lee, Y.J.: Hide-and-seek: Forcing a network to be meticulous for
  weakly-supervised object and action localization. In: 2017 IEEE International
  Conference on Computer Vision (ICCV). pp. 3544--3553. IEEE (2017)

\bibitem{szegedy2016rethinking}
Szegedy, C., Vanhoucke, V., Ioffe, S., Shlens, J., Wojna, Z.: Rethinking the
  inception architecture for computer vision. In: Proceedings of the IEEE
  Conference on Computer Vision and Pattern Recognition (CVPR). pp. 2818--2826
  (2016)

\bibitem{tompson2015efficient}
Tompson, J., Goroshin, R., Jain, A., LeCun, Y., Bregler, C.: Efficient object
  localization using convolutional networks. In: Proceedings of the IEEE
  Conference on Computer Vision and Pattern Recognition (CVPR). pp. 648--656
  (2015)

\bibitem{vaswani2017attention}
Vaswani, A., Shazeer, N., Parmar, N., Uszkoreit, J., Jones, L., Gomez, A.N.,
  Kaiser, {\L}., Polosukhin, I.: Attention is all you need. In: Advances in
  Neural Information Processing Systems (NIPS). pp. 5998--6008 (2017)

\bibitem{wah2011caltech}
Wah, C., Branson, S., Welinder, P., Perona, P., Belongie, S.: The caltech-ucsd
  birds-200-2011 dataset  (2011)

\bibitem{wan2013regularization}
Wan, L., Zeiler, M., Zhang, S., Le~Cun, Y., Fergus, R.: Regularization of
  neural networks using dropconnect. In: International Conference on Machine
  Learning (ICML). pp. 1058--1066 (2013)

\bibitem{wang2017residual}
Wang, F., Jiang, M., Qian, C., Yang, S., Li, C., Zhang, H., Wang, X., Tang, X.:
  Residual attention network for image classification. In: Proceedings of the
  IEEE Conference on Computer Vision and Pattern Recognition (CVPR). pp.
  3156--3164 (2017)

\bibitem{woo2018cbam}
Woo, S., Park, J., Lee, J.Y., So~Kweon, I.: Cbam: Convolutional block attention
  module. In: Proceedings of the European Conference on Computer Vision (ECCV).
  pp. 3--19 (2018)

\bibitem{xue2019danet}
Xue, H., Liu, C., Wan, F., Jiao, J., Ji, X., Ye, Q.: Danet: Divergent
  activation for weakly supervised object localization. In: Proceedings of the
  IEEE International Conference on Computer Vision (ICCV). pp. 6589--6598
  (2019)

\bibitem{yu2018generative}
Yu, J., Lin, Z., Yang, J., Shen, X., Lu, X., Huang, T.S.: Generative image
  inpainting with contextual attention. In: Proceedings of the IEEE Conference
  on Computer Vision and Pattern Recognition (CVPR). pp. 5505--5514 (2018)

\bibitem{zhang2018adversarial}
Zhang, X., Wei, Y., Feng, J., Yang, Y., Huang, T.S.: Adversarial complementary
  learning for weakly supervised object localization. In: Proceedings of the
  IEEE Conference on Computer Vision and Pattern Recognition (CVPR). pp.
  1325--1334 (2018)

\bibitem{zhang2018self}
Zhang, X., Wei, Y., Kang, G., Yang, Y., Huang, T.: Self-produced guidance for
  weakly-supervised object localization. In: Proceedings of the European
  Conference on Computer Vision (ECCV). pp. 597--613 (2018)

\bibitem{zhou2016learning}
Zhou, B., Khosla, A., Lapedriza, A., Oliva, A., Torralba, A.: Learning deep
  features for discriminative localization. In: Proceedings of the IEEE
  Conference on Computer Vision and Pattern Recognition (CVPR). pp. 2921--2929
  (2016)

\end{thebibliography}
\end{document}